# SCALE-Net: Scalable Vehicle Trajectory Prediction Network under Random Number of Interacting Vehicles via Edge-enhanced Graph Convolutional Neural Network*

Hyeongseok Jeon, *Student Member, IEEE*, Junwon Choi, and Dongsuk Kum, *Member, IEEE*

*Abstract*— Predicting the future trajectory of surrounding vehicles in a randomly varying traffic level is one of the most challenging problems in developing an autonomous vehicle. Since there is no pre-defined number of interacting vehicles participate in, the prediction network has to be scalable with respect to the vehicle number in order to guarantee the consistency in terms of both accuracy and computational load. In this paper, the first fully scalable trajectory prediction network, SCALE-Net, is proposed that can ensure both higher prediction performance and consistent computational load regardless of the number of surrounding vehicles. The SCALE-Net employs the Edge-enhance Graph Convolutional Neural Network (EGCN) for the inter-vehicular interaction embedding network. Since the proposed EGCN is inherently scalable with respect to the graph node (an agent in this study), the model can be operated independently from the total number of vehicles considered. We evaluated the scalability of the SCALE-Net on the publically available NGSIM datasets by comparing variations on computation time and prediction accuracy per single driving scene with respect to the varying vehicle number. The experimental test shows that both computation time and prediction performance of the SCALE-Net consistently outperform those of previous models regardless of the level of traffic complexities.

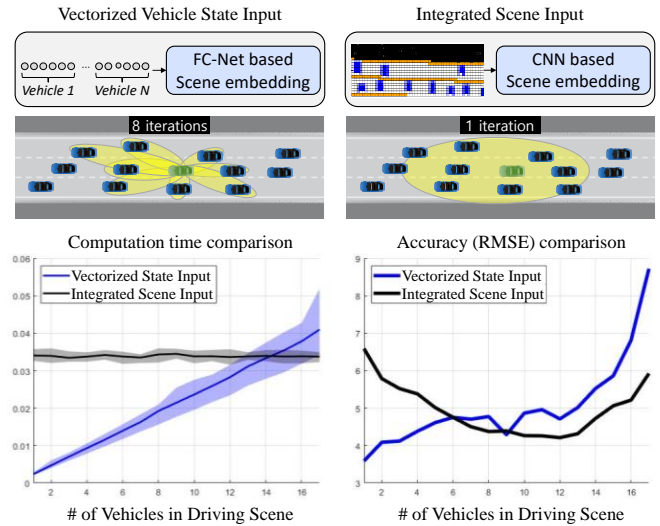

Figure 1. Comparison between state input based- and scene input based- prediction model on variation of computation time and accuracy per a single driving scene with respect to the number of surrounding vehicles

## I. INTRODUCTION

In order to succeed in self-driving tasks under the real-world situation, an autonomous vehicle has to cope with the various driving situation from the highway to the complex urban roads. Therefore, the consistency in both accuracy and inferencing time of the autonomous driving system in the various driving scene is important. Otherwise, an autonomous vehicle will be unstable under a randomly varying driving complexity

Among the key components of the autonomous vehicle system: perception, prediction, planning, and control, the prediction algorithm is the most difficult subsystem to guarantee the scalability because of the undefined inputs and outputs dimensions. The perception system is designed for the pre-defined sensor configuration to extract the states of multi-objects around ego vehicle. On the other hand, the planning and control systems produce a pre-defined number of output parameters, planned trajectory, and control input, using a multi- or single- input parameters, respectively. However, the prediction system has to be multi-input multi-output systems without any pre-defined dimensions of the parameters since the future of the random number of multiple surrounding vehicles is predicted using historical information of each vehicle. Therefore, ensuring the consistently fast computation time of the prediction networks with high performance under various driving complexity is a challenging problem.

However, in the previously developed future prediction models, there was no fully scalable high-performing model ensuring low computational loads. As shown in Fig. 1, the vectorized state input based prediction model, non-scalable model, shows remarkable performance in both computation time and accuracy only for the sparse driving situation. On the other hand, even though the integrated scene input based prediction model is fully scalable, overall performance is decreased in the less-complex driving situation with slow inference time because of blurred maneuver features of the individual entities in forming integrated scene architecture.

Therefore, to break through these trade-off effects, we propose a fully scalable network for future trajectory prediction of the surrounding vehicles, SCALE-Net, ensuring the consistent higher prediction performance with low computational power regardless of the traffic complexity. In SCALE-Net, EGCN-LSTM based sequential interaction

*This work was supported in part by the Technology Innovation Program under Grant 10083646 (Development of Deep Learning-Based Future Prediction and Risk Assessment Technology considering Inter-vehicular Interaction in Cut-in Scenario), funded by the Ministry of Trade, Industry, and Energy, South Korea.

H. Jeon is with the Cho Chun Shik Graduate School of Green Transportation, Korea Advanced Institute of Science and Technology (KAIST), Daejeon 34141, South Korea (e-mail: 521hsgugu@ kaist.ac.kr).

J. Choi is with the Department of Electrical Engineering, Hanyang University, Seoul 04763, South Korea (e-mail: junwchoi@hanyang.ac.kr).

D. Kum is with the Cho Chun Shik Graduate School of Green Transportation, Korea Advanced Institute of Science and Technology (KAIST), Daejeon 34141, South Korea (corresponding author to provide phone: +82-42-350-1266; fax: +82-42-350-1250; e-mail: dskum@ kaist.ac.kr).

embedding layer can inherently model the natures of inter-vehicular interaction using a novel graph-based driving scene representation framework. In particular, our model can be characterized as follows,

1. **Fully scalable future prediction model.** SCALE-Net is a fully scalable model in terms of the capable vehicle numbers. In other words, the model is completely flexible and can handle random traffic levels in various driving situations.

2. **Explicitly embedded the natures of interaction.** The proposed framework can inherently imitate the nature of the inter-vehicular interaction by exploiting the EGCN-LSTM interaction embedding layer.

## II. RELATED WORKS

The steps of the previous interaction-aware prediction frameworks can be classified into three categories based on the model scalability: 1) non-scalable prediction models, 2) fully scalable prediction models, and 3) partially scalable prediction models, as shown in Fig. 1, where partially scalable prediction models address both of state inputs and integrated scene inputs.

### A. Non-Scalable Prediction Models

The earliest approach to interaction-aware future trajectory prediction was to predict a single target vehicle considering its neighbors using time serial explicit states of the vehicles such as position and velocity. Therefore, only one vehicle is predicted in a single iteration.

In [1, 2], the future trajectory is generated using an LSTM based structure with time serial states of the target vehicle and ego vehicle. Another methodology reported in [3, 4] focused only on one-to-one interaction between target vehicle and ego vehicle. On the other hand, inter-vehicular interaction among multiple surrounding vehicles is considered in [5, 6]. In [5], the future trajectory is generated by LSTM in cooperation with a maneuver modality classification network. In [6], states of the surrounding vehicles are concatenated up to 7 vehicles in order to parameterize the Gaussian mixture model, which describes the distribution of future positions of a single target vehicle. Additionally, a Kalman neural network is proposed in [7] for interaction-aware trajectory prediction.

Because these models are focusing on the future trajectory of a single target vehicle, the ability to capture the context of the individual maneuver is remarkable due to explicitly exploited vehicle states. Additionally, because the networks use vectorized states input, it is computationally light in inferencing. However, because the models are not scalable, the computation time as well as the prediction error will increase as the number of vehicles increases.

### B. Fully Scalable Prediction Models

There were several approaches to interpret the traffic situation as a sequential image for future prediction algorithms. In these approaches, states of multiple vehicles are put into an integrated format. With this process, maneuver features of individual vehicles will be implicitly considered under integrated traffic scene representation frameworks.

One of the most popular methodologies is to construct an occupancy grid map, as stated in [8, 9]. In these papers, a dual-channel occupancy grid map is built using positions of vehicles and lane markings. In [8], future occupancy of each cell is predicted via CNN and LSTM. On the other hand, in [9], lane change intention of the surrounding vehicles is generated via CNN. In contrast with the occupancy grid map, in [10], to predict high-level traffic speed, sequential vehicle states are stacked to two-dimensional grid maps that consist of time and space dimensions. Another major branch for integrated driving scene based prediction is to use images. In [11, 12], semantic images from top view are used. In [11], future semantic images with uncertainty distribution are generated using CNN based architecture. Similarly, in [12], the encoder-decoder network with VGG-16 based encoder followed by one-shot or RNN based decoder is introduced. Additionally, the front-view semantic image is used as well in [13]. In [13], the future trajectory is generated via GRU based architecture using semantic scene context generated by CNN.

By using integrated driving scene input, networks can be fully scalable with respect to the number of vehicles because the dimensionality of the input shape is consistent even in a dense traffic situation. Additionally, the model is powerful to capture scene-level interaction contexts. Even though the model can assure the scalability, however, these approaches are computationally heavy for a single iteration and are not able to explicitly capture the characteristics of individual maneuvers.

### C. Partially Scalable Prediction Models

In [14, 15, 16, 17], hybrid approaches are introduced which are partially scalable by combining the strengths of the two approaches mentioned above. In [14, 15, 17], after individual historical trajectory encoding via LSTM encoder using explicit vehicle states, interaction among multiple agents is modeled by social pooling on the grid map. In [14], individually encoded human trajectories are pooled by occupancy map pooling. For vehicle applications, [15, 17], convolutional social pooling is applied to the context vectors of individual vehicles using a social tensor. On the other hand, in [16], state inputs are concatenated with a context vector generated by CNN using integrated scene input.

These approaches effectively combine the pros of both approaches. The partially scalable prediction models are scalable up to predefined vehicle numbers, usually 7 vehicles, and powerful to capture mid-level of inter-vehicular interaction.

However, in all of the previously developed algorithms, there was a trade-off between the model scalability and the ability to capture individual vehicle contexts, which affect individual future trajectory prediction accuracy. However, in the real driving scene, both of the scalability and interpretability of interaction contexts from individual vehicle level to overall traffic scene level are important. The SCALE-Net, proposed in this paper, is able to break through the trade-off between scalability and ability to interaction interpretation by introducing a novel state-based integrated scene representation framework with graphical modeling of the driving scene ensuring low computational load.

## III. OVERALL ARCHITECTURE

To overcome the trade-off effect appeared in the previous works, in this paper, a novel vehicle state-based integrated scene representation framework is proposed via graphical modeling of the driving scene, $G = (V, E)$, where graph ($G$), vertices (nodes) ($V$), and edges ($E$) stand for the interaction context of driving scene, the maneuver characteristics of individual vehicles, and a single one-to-one inter-vehicular interaction, respectively. Using a graphically modeled traffic scene, the first framework to construct an integrated driving scene based on explicit vehicle states is addressed. Using this driving scene representation method, the SCALE-Net can be fully scalable assuring the ability to interaction interpretation and low computation power with higher prediction accuracy.

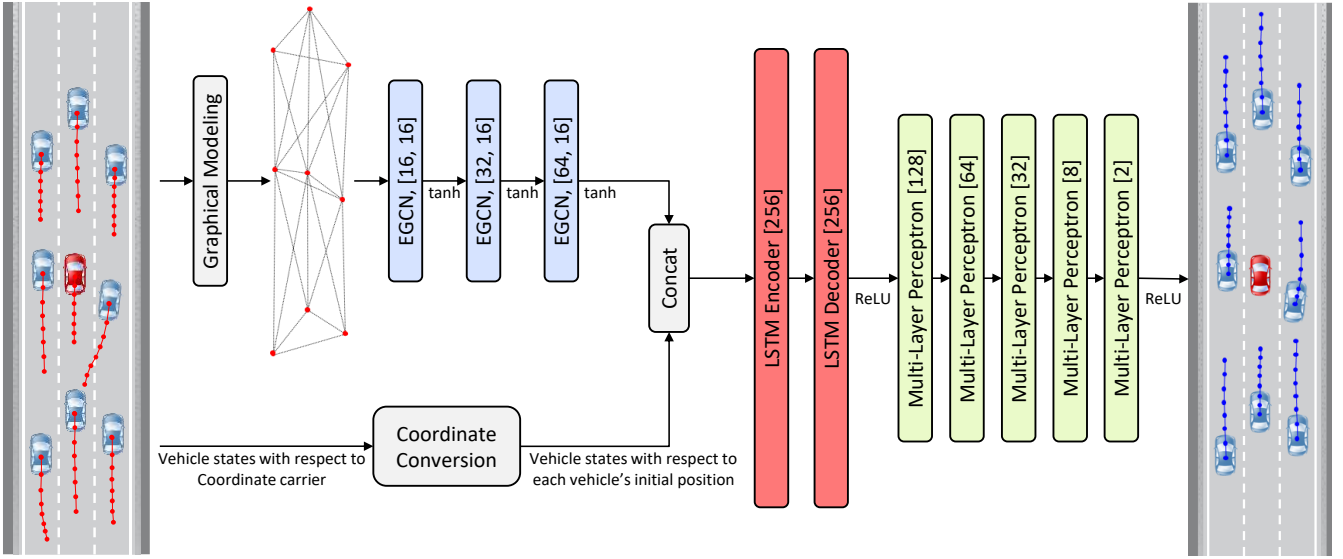

Figure 2. Overall architecture of the SCALE-Net for interactive scalable trajectory prediction algorithm. Historical states of the ego and surrounding vehicles, which is illustrated with dotted red line, is used as input parameter of the proposed architecture. After passing through the EGCN based scene embedding layer and LSTM based trajectory predictor, future trajectory of the surrounding vehicles are generated as shown in right-most figure with blue dotted line.

As shown in Fig. 2, the SCALE-Net is designed with EGCN-LSTM based interaction embedding layer and LSTM-MLP based future trajectory generation layer. Graphical modeling of the traffic scene was firstly introduced in [18]. However, the model proposed in [18] was not fully scalable, whereas the model proposed in this paper is fully scalable with various advantages stated above.

## IV. EGCN BASED INTERACTION EMBEDDING LAYER

In this paper, firstly, the traffic scenario is reconstructed via the graph on the constantly moving reference frame. Then, vehicle contexts are propagated to neighboring vehicles via EGCN operation. The concept of EGCN, a graph neural network variant, is introduced in [19]; EGCN updates the node feature via an attention mechanism induced by edge features. Since EGCN operation is close to the nature of the inter-vehicular interaction, EGCN based embedding framework in SCALE-Net can inherently interpret inter-vehicular interaction.

### A. Graphical Modeling of the Driving Scene

In order to exploit EGCN, the driving scene has to be modeled into a graph structure consisting of nodes and edges. In this paper, vehicles are treated as nodes in the graph. As shown in (1), the node feature matrix, $X$, consists of position $(x, y)$, velocity $(v_x, v_y)$, and heading angle $(\theta)$ of each vehicle, where $X_e$ and $X_k$ $[k = 1, 2, \cdots, n]$ represent states of the ego vehicle and of vehicle $k$. On the other hand, the edge feature matrix, $E$, consists of the absolute values of the relative states between two vehicles, formulated in (2) to (4), where $E \in \mathbb{R}^{(n+1) \times (n+1) \times 4}$.

$$\bar{X}_l = \begin{bmatrix} x_l & y_l & v_{x_l} & v_{y_l} \end{bmatrix} \quad (2)$$

$$\Delta \bar{X}_{lm} = \begin{cases} 0 & (if\ l = e) \\ |\bar{X}_m - \bar{X}_l| & (otherwise) \end{cases} \quad (3)$$

$$E = \begin{bmatrix} 0 & 0 & \cdots & 0 \\ \Delta \bar{X}_{1e} & 0 & \cdots & \Delta \bar{X}_{1n} \\ \vdots & \vdots & \ddots & \vdots \\ \Delta \bar{X}_{ne} & \Delta \bar{X}_{n1} & \cdots & 0 \end{bmatrix} \quad (4)$$

The physical meaning of the element of the edge feature matrix $E_{ij}$ is the interaction context from the vehicle $j$ to the vehicle $i$. Additionally, during forming the edge feature matrix, the direction of the connectivity between two vehicles is considered in the proposed framework. Excluding the ego vehicle, all of the connections of multiple vehicles are bi-directional. In the case of the ego vehicle, graphs from the surrounding vehicles to the ego vehicle are disconnected because the ego vehicle is a controllable agent, not to be predicted. In other words, interaction effects from surrounding vehicles on the ego vehicle do not need to be considered in the prediction algorithm. Therefore, the first row of the edge feature matrix becomes zero in (4) which stands for the interaction effect on the ego vehicle.

Finally, the diagonal elements of $E$ are modified from (4) based on (5) in order to equally weight importance between self-interaction, interaction effect on the vehicles from the own states, and sur-interactions, interaction effects on the vehicles from the surrounding vehicles, where $i = 1, 2, \cdots, n + 1$ and $k = 1, 2, 3, 4$.

$$X = \begin{bmatrix} X_e \\ X_1 \\ \vdots \\ X_n \end{bmatrix} = \begin{bmatrix} x_e & y_e & v_{x_e} & v_{y_e} & \theta_e \\ x_1 & y_1 & v_{x_1} & v_{y_1} & \theta_1 \\ & & \vdots & & \\ x_n & y_n & v_{x_n} & v_{y_n} & \theta_n \end{bmatrix} \quad (1)$$

$$E_{iik} = \begin{cases} \sum_{j=1}^{n+1} E_{ijk} & \left( \sum_{j=1}^{n+1} E_{ijk} \neq 0 \right) \\ 1 & \left( \sum_{j=1}^{n+1} E_{ijk} = 0 \right) \end{cases} \quad (5)$$

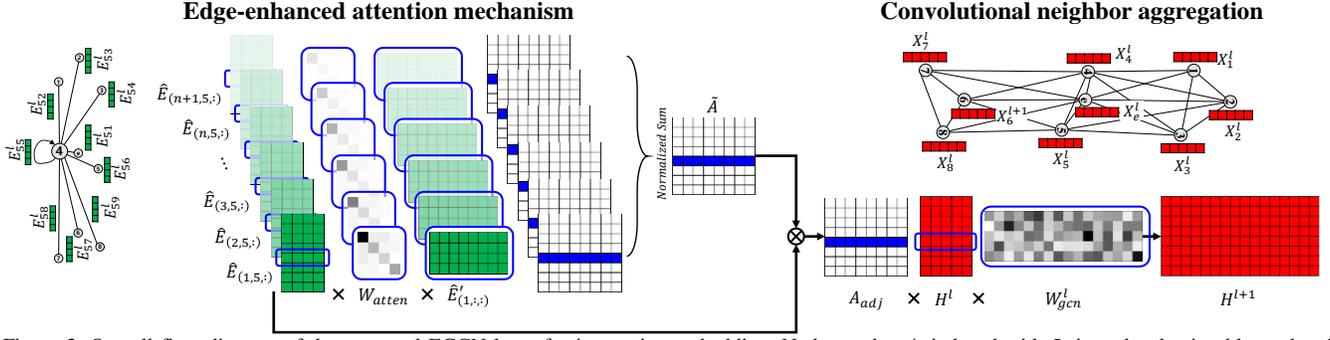

Figure 3. Overall flow diagram of the proposed EGCN layer for interaction embedding. Node number 4, indexed with 5, is updated using blue-colored elements. **Left**: in edge-enhanced attention process, weight of the vehicles around vehicle number 4 is calculated using the relative states of the entire vehicles in order to generated weighted adjacency matrix, $A_{adj}$. **Right**: using $A_{adj}$, node information of the vehicle 4 is updated by weight of GCN, $W_{gcn}$.

*B. Edge-enhanced Attention Mechanism*

There are two steps in EGCN operation, which one is an edge-enhanced attention mechanism and the other one is a graph convolutional neural network-based information update, as shown in Fig. 3. In order to update the context of the particular vehicle, firstly, the importance weights of each vehicle around the target vehicle have to be calculated. Even in real driving situations, when establishing a driving strategy, there are more possibly conflicting vehicles compared to others considering relative states of the vehicles. Therefore, to inherently imitate this feature, we propose an edge-enhanced attention mechanism. The goal of the edge-enhanced attention mechanism is to generate a weighted adjacency matrix, which defines the connectivity of the interaction pairs with connection strength.

First, the edge feature matrix has to be normalized vehicle-wisely in advance. Then, the attention matrix, $\tilde{A}$, is generated using the normalized edge feature matrix, $\hat{E}$, and the trainable attention weight, $W_{atten} \in \mathbb{R}^{4 \times f_{atten}}$, where $f_{atten}$ denotes the number of attention filters, as formulated in (6) and (7).

$$A = \sum_{l=1}^{n+1} \hat{E}_{l::} W_{atten} W_{atten}^T \hat{E}_{l::}^T \quad (6)$$

$$\tilde{A}_{ijk} = \frac{A_{ij}}{\sum_{l=1}^{n+1} A_{il}} \quad (7)$$

where $k$ represents the index of the edge features, which varies from 1 to 4 in our graph model and ":" in the subscript is used to select the entire range of the axis. The weighted adjacency matrix $A_{adj}$ is generated by element-wise multiplication between $\hat{E}$ and $\tilde{A}$.

*C. Feature Propagation via EGCN*

Using the node feature matrix and the weighted adjacency matrix, the effect of multiple interaction pairs simultaneously propagates over the whole traffic scene via the graph convolutional operation. Because of the neighbor-wise aggregation of a single interaction effect to the multi-vehicle situation, the operation is not affected by the total number of vehicles, a fully scalable model is realized. In the proposed EGCN based interaction embedding architecture, the hidden matrix consists of the hidden states of each vehicle; it is updated as follows:

$$H^l = \tanh\left(\sum_{k=1}^{4} A_{adj::k} H^{l-1} W_{gcn}^l\right), H^0 = X \quad (8)$$

where $H^l$ and $W_{gcn}^l \in \mathbb{R}^{f_{gcn}^{l-1} \times f_{gcn}^l}$ denote the hidden matrix and the trainable weight matrix of the $l^{th}$ EGCN layer, respectively. Additionally, $f_{gcn}^l$ represents the number of propagating filters in the $l^{th}$ EGCN layer, where $f_{gcn}^0 = 5$ in this paper. At the end of the $L^{th}$ EGCN layer, $H^L$ is generated; its rows represent the interaction effect of the whole traffic scene on vehicle $i$, $H_i^L$.

V. LSTM ENCODER-DECODER TRAJECTORY PREDICTOR

Time serial future positions of the surrounding vehicles are generated by conventional LSTM based seq2seq encoder-decoder model, which accepts the interaction-embedded hidden matrix $H^L$ as its input basis. By encoding inter-vehicular interaction sequentially via the LSTM encoder, time-varying interaction can be considered.

*A. Coordinate conversion*

In the trajectory predictor, each surrounding vehicle is placed in its own coordinate system in order to standardize the states of multiple vehicles for the generality of the model. Therefore, in the overall scenario, the origin of each vehicle's coordinate system is placed on the initial position of each vehicle. After exploiting the coordinate conversion operation, every row of the node feature matrix is re-defined forming the vehicle states matrix, $S$, under vehicle coordinates where $S_i$ is the state of vehicle $i$ on vehicle $i$'s coordinates.

*B. Seq2Seq Trajectory Predictor*

In the proposed trajectory predictor, the sequence-to-sequence (seq2seq) model with LSTM encoder-decoder architecture is used. The main function of the LSTM encoder is to encode sequential information of the interaction and vehicle states. The LSTM decoder serves as a future position generator for each surrounding vehicle considering explicitly embedded interaction effects of the driving scene on each vehicle.

In order to consider both individual maneuver context and interaction contexts, first, the input vector to LSTM encoder is established by concatenation operation between $S$ and $H_i^L$. Finally, time serial future positions of the vehicle $i$ is generated via the LSTM encoder-decoder model. In this paper, the historical time horizon, prediction time horizon, and time gap between adjacent time steps are designated as 3.5 seconds, 4 seconds, and 0.5 seconds, respectively.

## VI. EXPERIMENTAL EVALUATION RESULT

### A. Implementation

In the proposed SCALE-Net, there are three EGCN layers followed by a batch normalization layer. In each EGCN layer, there are two hyper-parameters, $[f_{gcn}, f_{atten}]$, where $f_{gcn}$ and $f_{atten}$ determines the dimension of the node feature in the next layer and the number of attention filters, respectively. After three EGCN layers are activated by $tanh$, the LSTM encoder and LSTM decoder are placed with 256-dimensional state and ReLU activation. Finally, future positions of the vehicles are generated by the Multi-Layered Perceptron (MLP) layer. All parameters for our model are shown in Fig. 2.

Throughout the training process, root mean squared error (RMSE) in meters was used as the loss function, as defined by (9). and (10).

$$dist_{ij}^k = \left(x_{ij}^k - \hat{x}_{ij}^k\right)^2 + \left(y_{ij}^k - \hat{y}_{ij}^k\right)^2 \quad (9)$$

$$RMSE = \sqrt{\frac{\sum_{i=1}^{N} \sum_{j=1}^{n_i} \sum_{k=1}^{8}(dist_{ij}^k)}{\sum_{i=1}^{N} n_i}} \quad (10)$$

where $dist_{ij}^k$ represents the squared distance error of the $j^{th}$ surrounding vehicle in the $i^{th}$ scenario at time step $k$.

Based on the RMSE loss function, using an ADAM optimizer, we trained the model twice with different learning and decay rates. First, the model was optimized with 0.0011 learning rate and 0.0001 decay rate for 100 epochs with 64 batch sizes. Both the learning rate and decay rate were set to 0.0001 for the second optimizer, for 50 epochs with identical batch size.

### B. Dataset

The performance of the SCALE-Net is evaluated on the NGSIM datasets [20, 21]. NGSIM dataset is a camera-recorded top-view video based naturalistic trajectory dataset recorded on US highway 101 (US-101) and interstate 80 (I-80). There are three temporally separated datasets for each recording site. Therefore, we designated the third set of data from I-80 as the test dataset. Since NGSIM data has a large number of lanes with dense traffic flow compare to another trajectory dataset, the HighD dataset [22], as shown in Fig. 4, there is a lot of interactive driving scene in NGSIM dataset compared to what is seen in the HighD dataset.

Since the naturalistic trajectory data is biased towards the lane-keeping scenario, most data were filtered out and only the balanced data is used with respect to the three maneuver categories of the ego vehicle: lane change left (LCL), lane change right (LCR), and lane keeping (LK) as shown in Table I. Additionally, to consider real-world situation, the range of the sensible boundary from the ego vehicle's point of view is

TABLE I. NUMBER OF DATA USED IN MODEL TRAINING

| Ego Maneuver | Training | Validation | Test | Total |
|---|---|---|---|---|
| LCL | 40,231 | 5,710 | 4,947 | 50,888 |
| LCR | 40,231 | 5,710 | 4,947 | 50,888 |
| LK | 40,231 | 5,710 | 4,947 | 50,888 |

limited up to 50 meters considering widely used LiDAR sensor in autonomous vehicle

### C. Scalability Evaluation

A fully scalable approach can ensure consistent computational time regardless of the vehicle number. Otherwise, because iterative computation is required to predict the entire vehicles, computational time increases according to the number of vehicles need to be predicted. Therefore, in this paper, the scalability of the model is evaluated using computing time required per a single driving scene under the various number of vehicles.

In the scalability test, the computational time of each model is measured 20 times and the average value with the fluctuation range is measured in identical hardware configuration. The scalability of the SCALE-Net is compared with three baselines as followings:

- *V-LSTM: vanilla-LSTM based non-scalable network for single vehicle prediction.*
- *CNN-LSTM: fully scalable prediction network via CNN based scene embedding using an occupancy grid map, as proposed in [8]*
- *CNN-FcNet: partially scalable prediction network described in [16] consisting of CNN and FcNet based scene embedding.*

As shown in Fig. 5, the SCALE-Net preserves 100 Hz of computational performance, regardless of the number of vehicles, whereas other approaches cannot. Even though the non-scalable network, V-LSTM, shows remarkable high computational efficiency for a single iteration, the total computation time for a single driving scene prediction gets higher as the more vehicles participate in the driving scene because of the necessity of the multiple iterations. In the previous fully scalable model, CNN-LSTM, consistency of the inferencing time is guaranteed as shown in Fig. 5. However, the computational load per single iteration is remarkably heavy. In other words, conventional fully scalable models for trajectory prediction can ensure the consistent computation time with low speed. In terms of CNN-FcNet, a partially scalable model, the total computation time per a single driving scene is step-wisely increased because of the limitation on the maximum capable vehicle number. However, in SCALE-Net, the consistent computation time for random traffic levels is realized by exploiting a novel state-based integrated scene representation framework.

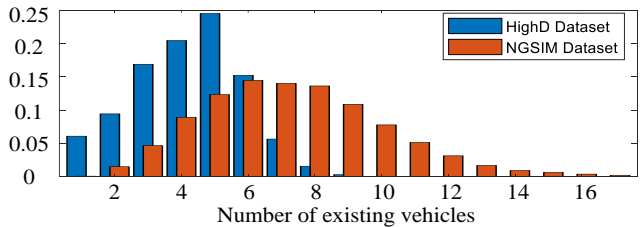

Figure 4. Normalized number of the NGSIM dataset and HighD dataset with respect to number of the vehicles in a single driving which indicates how many complex scene is contained in.

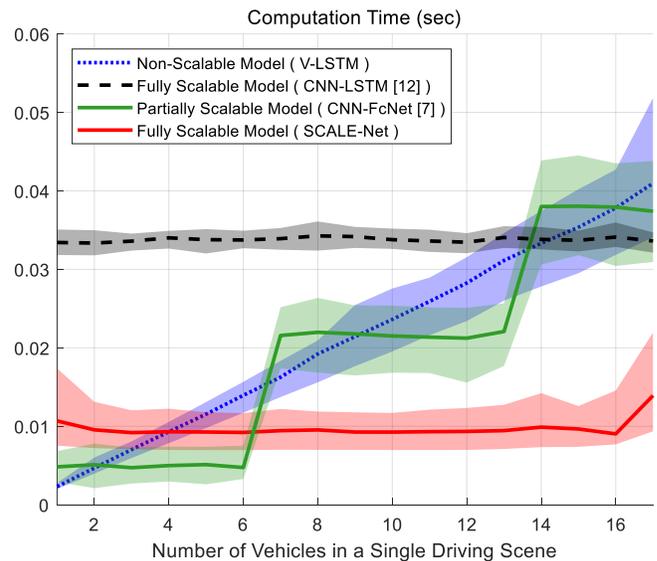

Figure 5. Scalability test result of the previously developed models and the proposed method, SCALE-Net, in terms of computation time per a single driving scene with various number of surrounding vehicles.

TABLE II. Prediction accuracy and performance enhancement ratio comparison with baselines

| Prediction Horizon | V-LSTM (CS-LSTM) | CS-LSTM | V-LSTM (MATF) | MATF | V-LSTM (Proposed) | SCALE-Net |
|---|---|---|---|---|---|---|
| 1 sec | 0.68 | 0.61 (10.3 %) | 0.66 | 0.66 (0 %) | 0.67 | **0.459 (31.5 %)** |
| 2 sec | 1.65 | 1.27 (23.0 %) | 1.62 | 1.34 (17.3 %) | 1.81 | **1.156 (36.1%)** |
| 3 sec | 2.91 | 2.09 (28.2 %) | 2.94 | 2.08 (29.3 %) | 3.17 | **1.973 (37.8%)** |
| 4 sec | 4.46 | 3.10 (30.5 %) | 4.63 | 2.97 (35.9 %) | 4.77 | **2.911 (39.0 %)** |

*D. Prediction Accuracy Evaluation*

As stated in the previous section, SCALE-Net has a remarkable ability to handle random traffic levels consistently. Additionally, our model has prediction accuracy enhancement in terms of RMSE because of the additional effect from the model scalability. Because the proposed fully scalable model is able to capture entire interactions in any complexities of road situation, whereas not fully scalable models can only partially observe the interactions, the ability to interpret inter-vehicular interaction is improved. To evaluate this strength of SCALE-Net, in this section, prediction accuracy is compared with SOTA baselines as follows:

- *CS-LSTM: LSTM predictor with interaction embedding via convolutional social pooling introduced in [15]*

- *V-LSTM (CS-LSTM): Vanilla LSTM, designed in [15]*

- *MATF: LSTM predictor with convolutional social pooling based interaction embedding using the additional top-view image, described in [17]*

- *V-LSTM (MATF): Vanilla LSTM, designed in [17]*

- *V-LSTM (Proposed): Vanilla LSTM, designed in this work*

Table II and Fig. 6 show overall performance comparison results calculated using RMSE measure on the NGSIM datasets. Additionally, for a fair comparison, the performance enhancement ratio from the identical baseline (V-LSTM) is calculated with a percentage value for all of the previous works. In complex and interactive situations, based on the NGSIM dataset, the performance improvement ratio increases with an enlarged prediction horizon, as shown in Table II. Because of long-lasting historical interaction leverage in dense traffic situations, the more accurate interaction-aware model is exploited, the higher performance is achievable throughout the entire prediction horizon. Additionally, the proposed model improved the short-term prediction accuracy which was not achieved in the previous works. For short-term prediction, prediction accuracy is directly affected by how well historical interaction is captured. Therefore, we claim that the proposed fully scalable future prediction network is excellent at inherently capturing the multiple inter-vehicular interactions as well as ensuring the generality of the model under the random traffic complexities.

Additionally, as shown in Fig. 7, we also present the accuracy fluctuation with respect to the increasing traffic complexity in terms of the number of vehicles in a single driving scene. From Fig. 7, we note that the state-input model, V-LSTM, shows higher accuracy compared to the scene-input based model in the sparse traffic situation. This implies that individual maneuver features are much more important to execute future motion than the interaction effect from the traffic scene in a sparse driving scene. On the other hand, in dense traffic scenarios, the scene-input based prediction model outperforms the state-input based model. Because the vehicles are actively interacting with each other, in the complicated driving scenes, the future maneuver of the particular vehicles are highly dependent on the surrounding vehicles. Therefore, as described in Fig. 7, the ability of interaction interpretation is highlighted under the complex traffic scene. Finally, we show that the graph-based SCALE-Net outperforms the previous approaches in any level of traffic density. Because the graph model used in the proposed network can inherently model the entire levels of inter-vehicular interaction from individual maneuver to traffic-level interactions, we claim that SCALE-Net is generally showing higher accuracy assuring low computational load and scalability.

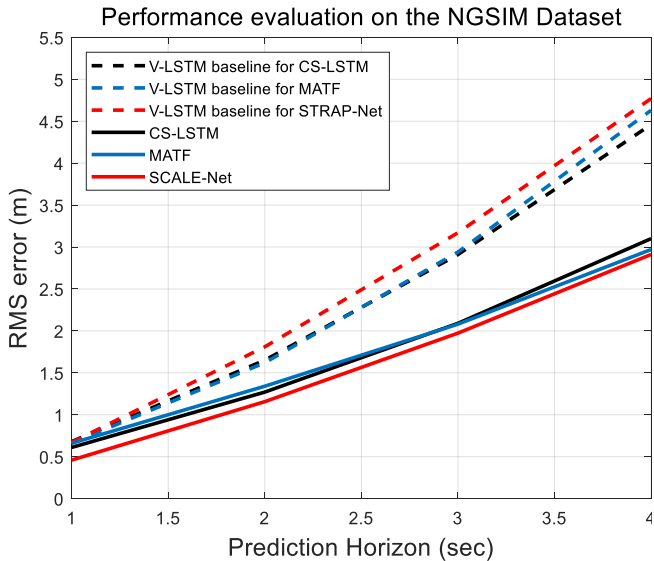

Figure 6. Accuracy comparison using RMSE measure and performance enhancement ratio of SCALE-Net compared to the baselines using NGSIM dataset.

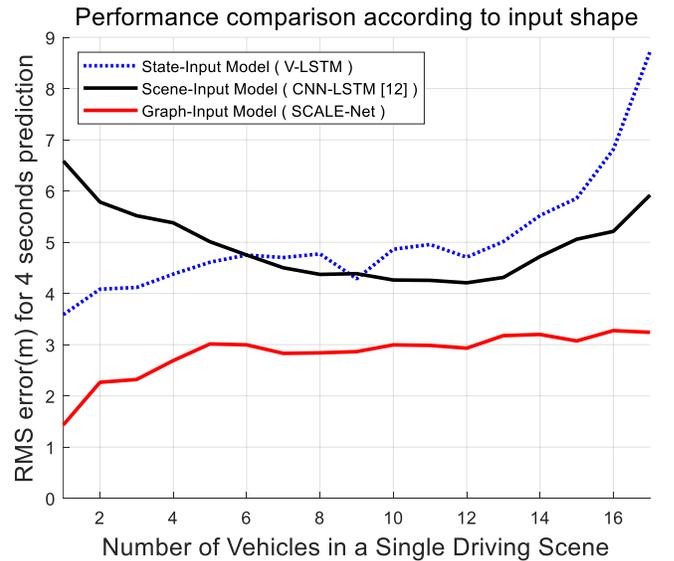

Figure 7. Changes in 4 seconds prediction accuracy of the three models accepting different formats of input parameters, which determines the model scalability, computational load, and prediction performance.

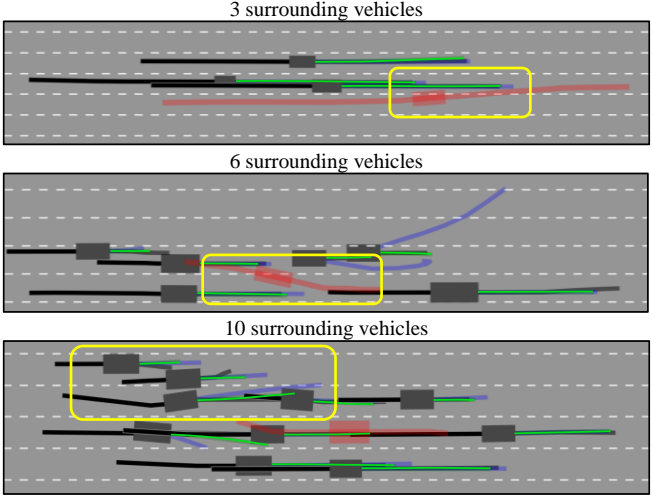

Figure 8. Examples of trajectory prediction result in various traffic level where green and transparent blue line is predicted by SCALE-Net and V-LSTM, respectively.

*E. Case Studies in a Specific Situation*

Representative scenarios for the case study were selected from three different traffic levels and the predicted trajectory from the V-LSTM baseline and SCALE-Net is compared, as shown in Fig. 8. Using the historical trajectory, illustrated as a solid black line, both models generate future trajectories of target vehicles around the red-colored ego vehicle, with transparent blue for V-LSTM and green solid line for SCALE-Net. The predicted future trajectory is compared with the ground truth of future trajectory as a transparent black line.

There is a highly interactive situation in the yellow box in Fig. 8. As traffic becomes more complex, fluctuating results are generated by V-LSTM, whereas SCALE-Net provides robust and accurate prediction results even in high-traffic situations because of its scalability. In addition, we observed that our model generated future trajectories that pushed away from each other, away from dangerously close surrounding vehicles, as shown in Fig. 9. Because of the risky interaction among vehicles 1 to 3, our model predicts the future trajectory of each vehicle by considering relational repulsive forces and gets closer to the ground truth than does the V-LSTM model. Our model interactively predicts longitudinal interaction between vehicle 1 and vehicle 2, whereas V-LSTM cannot capture the interactive braking maneuvers of vehicle 1. Additionally, SCALE-Net can capture lateral interaction between vehicle 3 and vehicle 2. Interaction effects from vehicle 2 to vehicle 3 make vehicle 3 waive its lane-change maneuver. Therefore, we note that by designing a fully scalable model in terms of the capable vehicle numbers, the ability to interpret the inter-vehicular interactions is improved as well. This is because SCALE-Net is able to capture the entire interaction in a single driving scene even in extremely complex situations.

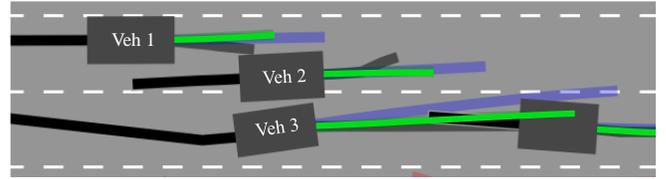

Figure 9. Critically interacting scene where the maneuver of the vehicles highly depends on interaction effect from adjacent vehicles.

## VII. CONCLUSIONS AND FUTURE WORK

Before this work, most of the researches on the deep learning architecture for future trajectory prediction focused on prediction accuracy. However, especially in real-vehicle applications, the scalability of the model is important as much as the prediction accuracy in order to cope with random traffic density in the real driving situation. According to the analytic studies in this paper, however, there was a trade-off between scalability and computational loads as well as between scalability and prediction accuracy. The higher scalability is realized, the higher computational resources are required. In addition, the previously developed prediction model with higher scalability is not enough to capture the context of the individual vehicle's maneuver feature because of its input shape which degrades prediction performance in the low traffic situations.

However, in this paper, the first fully scalable trajectory prediction model is addressed that can be applied to any random levels of traffic complexities using low computational resources. Additionally, the proposed SCALE-Net shows improved performance in prediction accuracy as well because the proposed model can predict the future trajectory of the target vehicle considering the entire interactions existing in the driving scene. Furthermore, by exploiting EGCN based interaction embedding and LSTM based sequential embedding, the natures of the inter-vehicular interaction, simultaneous propagative time-varying interaction, can be interpreted inherently. In particular, SCALE-Net can be characterized as follows:

- Fully scalable architecture with respect to vehicle numbers in a single driving scene so that the model is robustly applicable to any level of traffic density.

- The trade-off among the scalability, computation time, and performance is broken through via SCALE-Net.

- Inherently analyzed natures of the inter-vehicular interaction via EGCN based interaction embedding.

However, SCALE-Net cannot consider the road structures, one of the most prominent features of the traffic scene.

Because the infrastructural cues affect vehicle maneuvers a lot, a future prediction framework that incorporates infrastructural information has to be developed. The infra-aware prediction model with scalability will ensure the generality of the model in terms of the various road and/or infrastructure, whereas SCALE-Net has the generality in terms of the traffic complexity.